\documentclass[a4, 10pt, conference]{IEEEtran}
\IEEEoverridecommandlockouts

\usepackage{cite}
\usepackage{amsmath,amssymb,amsfonts}
\usepackage{amssymb,graphicx,xcolor,acronym,algorithm,url}
\usepackage[noend]{algpseudocode}
\usepackage{graphicx}
\usepackage{textcomp}
\usepackage{xcolor} 

\usepackage[export]{adjustbox}
\usepackage{color}

\usepackage{multirow}
\usepackage{subcaption}
\usepackage{hyperref}
\usepackage{bm}

\usepackage{amsthm}

\theoremstyle{definition} % plain style, no italics, no bold
\newtheorem{definition}{Definition}

\begin{document}

\title{SSR: Sparse Segment Reduction for Ternary GEMM Acceleration}

\author{
    \IEEEauthorblockN{Adeline Pittet, Shien Zhu, Valérie Verdan, Gustavo Alonso}
    \IEEEauthorblockA{ Systems Group, ETH Zurich, Switzerland \\
         apittet@ethz.ch, shien.zhu@inf.ethz.ch, verdanv@student.ethz.ch, alonso@inf.ethz.ch
    }
    
%     \IEEEauthorblockN{Adeline Pittet}
%     \IEEEauthorblockA{Systems Group, ETH Zurich \\
%         Zurich, Switzerland \\
%         apittet@ethz.ch
%     }
% \and
%     \IEEEauthorblockN{Shien Zhu}
%     \IEEEauthorblockA{Systems Group, ETH Zurich \\
%         Zurich, Switzerland \\
%         shien.zhu@inf.ethz.ch
%     }
% \and
%     \IEEEauthorblockN{Valérie Verdan}
%     \IEEEauthorblockA{Systems Group, ETH Zurich \\
%         Zurich, Switzerland \\
%         verdanv@student.ethz.ch
%     }
% \and
%     \IEEEauthorblockN{Gustavo Alonso}
%     \IEEEauthorblockA{Systems Group, ETH Zurich \\
%         Zurich, Switzerland \\
%         alonso@inf.ethz.ch
%     }
}

\IEEEaftertitletext{\vspace{-1.5\baselineskip}}

\maketitle

\begingroup
\renewcommand{\thefootnote}{}
\footnotetext{%
\footnotesize
\copyright~2026 EDAA. This is the author's version of the work.
It is posted here for personal use. The definitive version was
published in the Proceedings of the Design, Automation \& Test
in Europe Conference (DATE 2026),
DOI: 10.23919/DATE69613.2026.11539457.%
}
\endgroup

\begin{abstract} 
%In addition, Ternary LLMs further reduce the computation cost by replacing the multiplication operations with addition and subtraction. 
%However, since traditional sparse formats are designed for general scientific computing, they cannot exploit the structural properties of ternary networks. Although related works like BitNet and RSR++ are optimized for ternary LLMs, they cannot fully utilize the ternary feature to reduce the computation complexity and memory footprint.
%However, existing ternary-optimized methods like RSR++ do not exploit sparsity, while traditional sparse formats cannot fully utilize the ternary structure, leading to a gap of algorithms making use of both sparsity and ternary structure.

Large Language Models (LLMs) require substantial computational resources, limiting their deployment on resource-constrained hardware. Ternary LLMs mitigate these demands through weight quantization via ternary values, achieving significant compression often with $\mathbf{50-90\%}$ sparsity. However, existing approaches have limitations: methods optimized for ternary weights, such as BitNet, redundant segment reduction (RSR), and its improved version RSR++, do not exploit sparsity structures, while conventional sparse formats neglect ternary characteristics, foregoing dual optimization opportunities.

In this paper, we introduce \emph{Sparse Segment Reduction (SSR)}, a ternary matrix multiplication method designed to accelerate the inference of ternary LLMs and general Ternary Weight Networks (TWNs). SSR has a dedicated optimized ternary data format and an algorithm that systematically exploits sparsity patterns through computation trees that scale with the sparsity. SSR provides theoretical gains with asymptotically faster inference than RSR++ for sparsity above 50\%, while practical evaluations reveal performance improvements across all sparsity levels. Evaluation results show that SSR achieves 2.1-11.3$\times$ speedup over RSR++ on ternary GEMM with 45-95\% sparsity. Furthermore, SSR achieves 3.5-6.3$\times$ end-to-end speedup and 4.9\% of memory saving over RSR++ on the Llama-3 1B model inference. % Our single-threaded SSR implementation accelerates ternary MLP inference by \textbf{15.2$\times$ to 69$\times$} compared to RSR++ across all sparsity levels.

\end{abstract}


\section{Introduction}
% 1. LLM to TWN

% Driven by recent advances in deep learning, Large Language Models (LLMs) have emerged as powerful tools across a variety of domains, including recommendation systems \cite{Raza2025}, financial analysis \cite{Lee_2025}, autonomous cars \cite{10.1145/3180155.3180220} and healthcare diagnostics \cite{Cascella2024, AlSaad2024}.
Large Language Models (LLMs) have emerged as a powerful tool across many domains, yet their substantial computational and memory requirements limit their usage on resource-constrained devices \cite{seymour2024largelanguagemodelssmall}. Quantization addresses this \cite{MLSYS2024_42a452cb, kong2024tinyagent} by reducing bit precision of weights and activations, decreasing the memory usage, the energy consumption, and the inference time while aiming to close the accuracy gap. For example, Nvidia and AMD GPUs adopt FP8, FP6, and even FP4 quantization for LLM inference.

%%In practice, many CPU and GPU deployments adopt low-precision formats such as BFLOAT16 \cite{BFLOAT16_ArXiv_2019}, as well as 8-bit, and 4-bit integer encodings \cite{INT8_Nvidia_Arxiv_2020, UINT8_CVPR_2020, DA-INT8_AAAI_2021, NIPS_2022_FP8, ECCV_2024_INT4, NIPS_2023_INT4, Zhang_2023}. %gleich wie du es bisher hattest 
\emph{Binary Weight Networks (BWNs)} constrain weights to $\{-1,+1\}$ for extreme compression, which introduces an accuracy drop and requires dense computation~\cite{NIPS2015_3e15cc11}. \emph{Ternary Weight Networks (TWNs)} expand the weight space to $\{-1,0,+1\}$, enabling higher accuracy and efficiency while introducing sparsity~\cite{li2022ternaryweightnetworks, zhu2016trained}. Compared with FP8 quantization, TWNs achieve a $4\times$ compression and replace the multiplication operations in GEMM with lightweight addition and subtraction operations~\cite{li2022ternaryweightnetworks, Razani_2021_CVPR}. \emph{Ternary-Weight LLMs} apply TWNs to large language models, reducing size and accelerating inference~\cite{ma2025era}. %Ternary-Weight LLMs have been proposed to further reduce the model size and computation complexity \cite{ma2025era}. \emph{Binary Weight Networks (BWNs)} constrain the weights to $\{-1, +1\}$ to achieve extreme compression, but BWNs introduce obvious accuracy drop and requires dense computation \cite{NIPS2015_3e15cc11}. \emph{Ternary Weight Networks (TWNs)} extend the weight space to $\{-1, 0, +1\}$ and significantly outperform BWN in model accuracy and computation efficiency thanks to the weight sparsity ~\cite{li2022ternaryweightnetworks, zhu2016trained}.  
Ternary-Weight LLMs inherit the features of TWNs in the LLM era and share the same addition-based sparse GEMM on ternary weights and dense activations.

% sparsity over 50%: Gap to further optimization
Ternary LLM and general TWNs exhibit high sparsity which frequently exceeds $50\%$ and even surpasses $90\%$ in practice~\cite{DENG2022221}. For example, induced weight sparsity by learning binary gate variables reaches $95.84\%$~\cite{Srinivas_2017_CVPR_Workshops}, Faraone et al. achieve $97.6\%$ sparsity in ternary networks using quantization threshold regularization and pruning~\cite{10.1007/978-3-319-70096-0_41}, and up to $99.58\%$ has been reported in recent work~\cite{DENG2022221}. Since zero-valued weights contribute nothing to matrix multiplications, this inherent sparsity provides significant opportunities for computational and memory optimizations. Conservative methods typically target around $50\%$ sparsity for balanced trade-offs~\cite{8297304, zhu2017trainedternaryquantization}, whereas in practice $90$--$95\%$ sparsity is achievable without notable accuracy loss~\cite{DENG2022221}. 

However, the ternary-weight dense-activation sparse GEMM (ternary GEMM) is under-optimized. Traditional sparse formats such as CSR, JDS, and COO~\cite{tensorflow_csr_2020, MONTAGNE200487, nvidia_sparse_formats_2024} are designed for general scientific computing and cannot exploit the structural properties of ternary networks. Similarly, general-purpose libraries such as Eigen and PyTorch Sparse provide optimized sparse GEMM kernels but do not exploit the ternary structure of TWNs. Consequently, existing inference frameworks fail to translate the given structural properties into meaningful computational benefits. For example in our experiments, PyTorch Sparse GEMM based on CSC is slower than PyTorch Dense GEMM with 90\% or lower sparsity on CPUs, which is much lower than the 10$\times$ theoretical speedup of sparse GEMM. Though related works have proposed algorithms that use the ternary structure such as BitNet.cpp \cite{ArXiv_2025_BitNet.cpp}, RSR, and RSR++~\cite{RSR_ArXiv_2025}, they do not fully exploit the sparsity feature of ternary weights to reduce the computation complexity. In addition, their data formats have redundant values that brings larger memory footprint in ternary GEMM.  

% 6. In this paper....
In this paper, we propose \emph{Sparse Segment Reduction (SSR)} to achieve efficient ternary GEMM for ternary LLMs and general TWNs. We achieve asymptotic improvements over BitNet \cite{ArXiv_2025_BitNet.cpp} and RSR++~\cite{RSR_ArXiv_2025} in ternary GEMM independent of hardware or structured sparsity constraints. SSR systematically leverages the inherent sparsity patterns of TWNs through preprocessing stages that exclude zero-pattern elements, yielding compact computation trees that scale well with sparsity. During inference, these optimizations eliminate redundant operations and reduce memory traffic, achieving asymptotically superior performance for matrices with sparsity levels exceeding $50\%$, while delivering practical performance gains across low and high sparsity levels compared to RSR++~\cite{RSR_ArXiv_2025}.

% 7, show some key benchmarking results. 
%\textcolor{red}{2. BENCHMARK our popcount-optimized algorithm with an ablation study against related works.]}

We implement the SSR algorithm with code optimizations like unrolling, AVX2 SIMD, and multi-threading on CPUs (code is available at \cite{github-link}). Evaluation results show that SSR achieves 2.1-11.3$\times$ speedup over RSR++ on ternary GEMM with 45-95\% sparsity. SSR also achieves up to 59.4\% faster speed over PyTorch Sparse GEMM. SSR achieves 1.8-7.5$\times$ speedup up to 17.8\% of memory saving over RSR++ at the layer level on Llama-MLP blocks. Furthermore, SSR achieves 3.5-6.3$\times$ end-to-end speedup and 4.9\% memory saving over RSR++ on Llama-3 1B model inference.

\section{Background and Related Works}

\subsection{Target Problem}
We target ternary LLMs and general Ternary-Weight Deep Neural Networks, and Binary Weight Networks (BWNs) can be seen as a special case of TWNs. Let $X \in \mathbb{R}^{M \times K}$ denote a dense activation matrix and $W \in \{-1, 0, 1\}^{K \times N}$ a ternary weight matrix with a given sparsity level $p$. The primary computation in these networks involves the matrix product
\[
Y = X \cdot W,
\]
which forms the core of the inference workload. We represent the elements, rows, and columns of a weight matrix $W$ using $W[i,j]$, $W[i,:]$, and $W[:,j]$, respectively.  
And $\vec{x}_j$ is the $j$-th row of a matrix $X$.   
We denote the sparsity level of a matrix by $p$, representing the fraction of zero weights. Let $\sigma_n$ be a permutation of $n$ elements in $[n]$, i.e., a bijective mapping.   
Finally, let $\texttt{pattern}_0$ be the random variable counting the number of rows in a matrix that match the all-zero bit pattern.

\subsection{Quantized Neural Networks}
Quantization of neural networks has evolved from binary weights~\cite{NIPS2015_3e15cc11, 8103902, huang2019efficient} to ternary weights~\cite{li2022ternaryweightnetworks, DENG2022221, ardakani2019learningrecurrentbinaryternaryweights}, with recent work extending these approaches to large language models~\cite{ma2025era}. Advances also include hardware-friendly implementations~\cite{2017HW} and new training methods for ternary networks~\cite{NEURIPS2024_b8f10193}. Hardware-optimized solutions for quantized networks include specialized accelerators using low-bit arithmetic~\cite{guo2025fastmatrixmultiplicationslookup, T-MAC_MM_2021, 10.1145/3287624.3287721}, configurable FPGAs~\cite{10.1145/3270764}, and symmetry-aware in-memory computing~\cite{8226999}. Optimizations for general-purpose platforms leverage ARM NEON SIMD~\cite{trusov2022fastmatrixmultiplicationbinary}, GPUs~\cite{10.1145/3410463.3414654}, and GPU Tensor Cores for sparse ternary operations~\cite{10817836}.

%Quantized neural networks have motivated a wide range of hardware-optimized solutions. These include specialized accelerators leveraging low-bit arithmetic~\cite{guo2025fastmatrixmultiplicationslookup, T-MAC_MM_2021, 10.1145/3287624.3287721}, configurable FPGA designs with custom adder trees~\cite{10.1145/3270764}, and symmetry-aware in-memory computing architectures~\cite{8226999}. Further optimizations target general-purpose platforms, such as ARM CPUs with NEON SIMD~\cite{trusov2022fastmatrixmultiplicationbinary}, GPUs~\cite{10.1145/3410463.3414654}, and GPU Tensor Cores designed for sparse ternary operations~\cite{10817836}. 

%Quantization of neural networks has progressed from early binary weight representations~\cite{NIPS2015_3e15cc11, 8103902, huang2019efficient} to ternary weights \cite{li2022ternaryweightnetworks, DENG2022221, ardakani2019learningrecurrentbinaryternaryweights}, with recent developments extending these ideas to large language models~\cite{ma2025era}. Researchers are also exploring hardware-friendly implementations~\cite{2017HW} for structured ternary coding and new training methods for ternary networks~\cite{NEURIPS2024_b8f10193}. 
\subsection{Ternary LLM Inference}
%\todo{Introduce the drawbacks of RSR and RSR++.}

Ternary LLMs such as BitNet b1.58~\cite{ma2024era1bitllmslarge} achieve full-precision performance using ternary weights $\{-1,0,1\}$, offering a 2.71× speedup and 3.55× memory reduction, demonstrating the potential of extreme quantization for large language models. Current state-of-the-art ternary GEMM methods include T-MAC \cite{T-MAC_MM_2021}, which uses lookup tables to achieve 4–5× speedups via precomputed partial sums, and RSR++~\cite{RSR_ArXiv_2025}, which achieves logarithmic theoretical complexity improvement through redundant segment reduction. However, RSR++ faces significant practical limitations: it ignores weight sparsity by treating zeros as non-zero weights, transfers large amounts of unnecessary data via dense permutation arrays, performs many computations on elements that do not affect the output, and causes non-linear memory access patterns that degrade cache performance and increase computational overhead.

%\section{Preliminaries} 

%\subsection{Notation}
% We denote matrices by uppercase letters and vectors by lowercase letters. 

%\subsection{Problem Setting}
%We consider quantized deep neural networks (DNNs) with binary and ternary weight matrices, where the inference bottleneck typically lies in sequential activation-to-weight matrix multiplications. Binary Weight Networks (BWNs) can be seen as a special case of Ternary Weight Networks (TWNs).

\section{Method} 
\begin{figure*}[t]
    \centering
    \hspace{5cm}
    \includegraphics[width=0.9\textwidth]{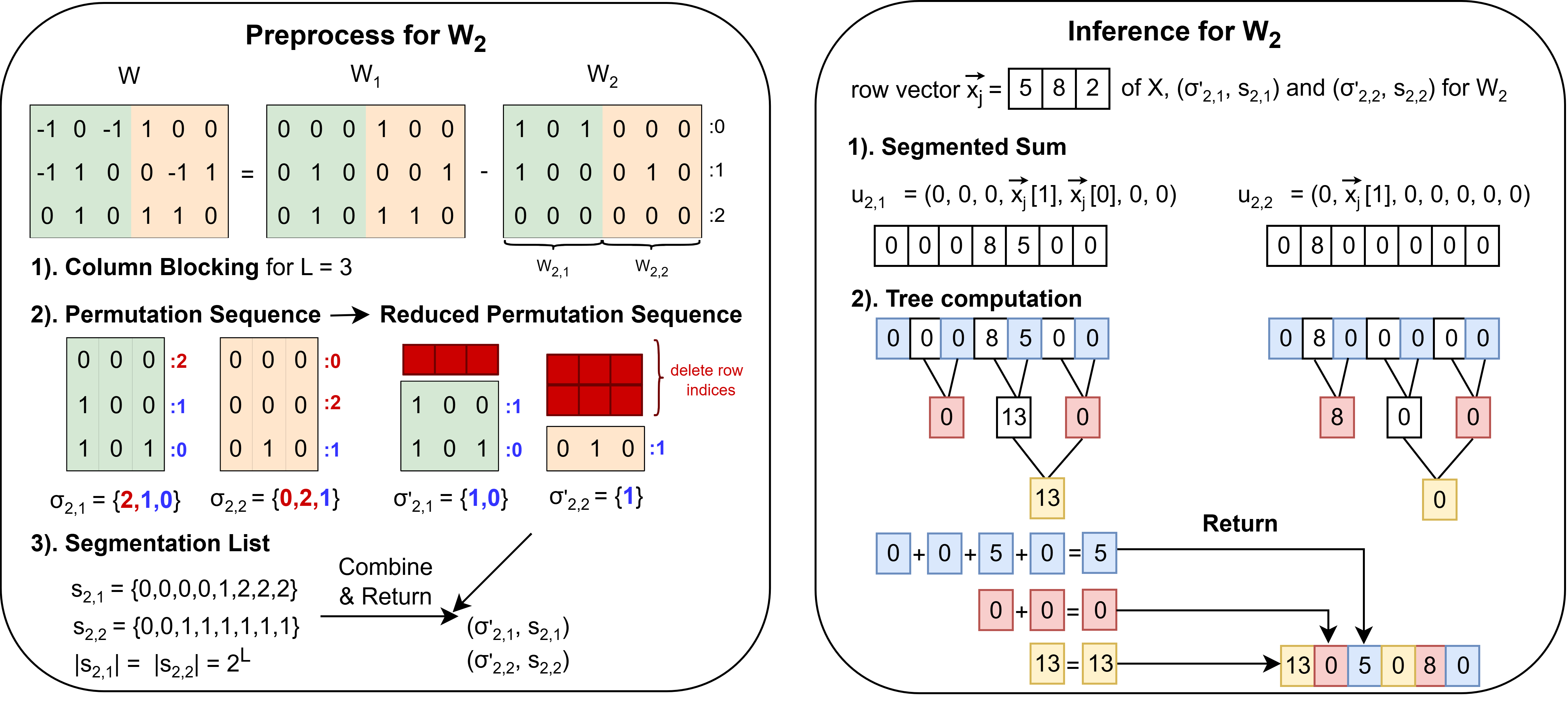}
    \caption{Pre-processing of the binary matrix $W_2$ and inference with the row vector $\vec{x}_j$ of the input matrix $X$ for $L=3$. The length of the reduced permutation sequence equals the number of nonzero patterns in the column blocks: $2$ for $W_{2,1}$ and $1$ for $W_{2,2}$. Row indices of all-zero rows are deleted and shown in red, while the remaining row indices are shown in blue.}
    \label{fig:example}
\end{figure*}

We introduce SSR, a new sparsity-aware method inspired by RSR++ that constructs block-wise indices excluding zero-pattern elements during preprocessing, enabling compact computation trees. During inference, these tree indices provide direct access to non-zero elements, avoiding redundant computations. Both stages are shown in Fig.~\ref{fig:example}. 

\subsection{Pre-processing}

Given a weight matrix \( W \in \{-1, 0, +1\}^{K \times N} \), we decompose $W$ into two binary matrices \( W_1 \) and \( W_2 \) such that $
W = W_1 - W_2$, where \( W_1[i,j] = 1 \) if \( W[i,j] = +1 \), otherwise \( 0 \); similarly, \( W_2[i,j] = 1 \) if \( W[i,j] = -1 \), otherwise \( 0 \). Then \( W_1 \) and \( W_2 \) go through the following preprocessing steps. See Fig. \ref{fig:example} for an example with $W = W_1 - W_2$.

\subsubsection{Column Blocking} 

First, we partition a binary matrix into consecutive column blocks of $L$ columns each. The last block may contain fewer columns in certain cases.

\begin{definition}[$L$-Column Block]
Given a binary matrix \( W \in \{0,1\}^{K \times N} \) and a blocking factor \( L \in \mathbb{N} \), partition \( W \) into column blocks \( W_i^{[L]} \) for \( i \in \{1, \dots, \lceil N/L \rceil\} \), where each block \( W_i^{[L]} \) consists of columns $W[:,(i-1) \cdot L]$ to $W[:, \min (i\cdot L-1, N-1)]$. 
\end{definition}

If \( L \) is clear from the context, we simply write \( W_i \) instead of \( W_i^{[L]} \) to denote a $L$-column block. In Fig. \ref{fig:example} with $L=3$, we partition $W_2$ into the column blocks $W_{2,1}$ and $W_{2,2}$.

\subsubsection{Permutation} 

To efficiently aggregate grouped binary patterns, we process the column blocks \( W_i \) of a binary matrix according to a permutation order that sorts the binary patterns in lexicographic ascending order.

\begin{definition}[Permutation Sequence]
Let \( W_i \in \{0,1\}^{K \times L} \) be a binary matrix. For each row $r \in \{0, \dots , K-1 \}$ of \( W_i \) define its \emph{key value} \( \kappa_i(r) \) as the integer value obtained by taking the concatenation of \( W_i[r,0], \dots, W_i[r,L-1] \). The \emph{permutation sequence} of \( W_i \), denoted by \( \sigma_i \), is a permutation of \( \{0, \dots, L-1\} \) such that
\[
\kappa_i(\sigma_i(0)) \leq \kappa_i(\sigma_i(1)) \leq \dots \leq \kappa_i(\sigma_i(L-1)).
\]
\end{definition}
Essentially, the permutation sequence \( \sigma_i \) for a column block \( W_i \) specifies how the rows of \( W_i \) should be reorganized so that their binary sequences, read as integer values, are arranged in ascending order. In Fig.~\ref{fig:example}, $W_{2,1}$ has binary patterns 000, 001, and 101 from rows 2, 1, and 0. Their ascending order ($000 < 001 < 101$) gives the row permutation $\sigma_{2,1}=\{2,1,0\}$.

\begin{definition}[Reduced Permutation Sequence]
Given a permutation sequence \(\sigma_i\) of a column block \(W_i\), the \emph{Reduced Permutation Sequence} \(\sigma_i'\) is obtained by removing a prefix of \(\sigma_i\). Formally, let \(q\) be the number of all-zero rows in \(W_i\). For all \(j = 0, \ldots, q-1\), we have
\[
\kappa_i(\sigma_i(j)) = 0,
\]
and \(\sigma_i'\) is the subsequence of \(\sigma_i\) obtained by excluding these first \(q\) indices.
\end{definition}

For the example in Fig.~\ref{fig:example}, $W_{2,1}$ has the permutation sequence $\sigma_{2,1} = \{2,1,0\}$. Since the matrix contains a single row with the all-zero pattern (row 2), this index can be removed to form the reduced permutation $\sigma'_{2,1} = \{1,0\}$.

\subsubsection{Segmentation}

To facilitate efficient computation on grouped binary patterns, we segment the sorted column blocks $W_i$ using boundary indices that mark transitions between different binary keys.

\begin{definition}[Segmentation List]
Let \(W_i\) be a column block with reduced permutation sequence \(\sigma_i'\), and let \(\pi_{\sigma_i'}(W_i)\) be the permuted block. The \emph{Segmentation Sequence} \(s_i\) consists of $2^L$ bounding indices of the interval containing the pattern
\[
s_i = \{b_0, \dots, b_{2^L-1}\},
\]
such that, in the sorted smaller matrix $\pi_{\sigma_i'}(W_i)$, for all $ j \in \{0, \dots, 2^L - 2\} \text{ and } r \in [b_j, b_{j+1}), \; \text{we have } \kappa_i(r) = j+1$, as we do not consider the all-zero pattern. 
\end{definition}

A segment $s$ encodes binary patterns as cumulative counts, where consecutive differences give the number of occurrences of each pattern.
For $W_{2,1}$ in Fig.~\ref{fig:example}, ignoring the all-zero pattern 000, we get $s = [0,0,0,0,1,2,2,2]$ representing cumulative counts for the bit pattern 001 to 111: the first four positions remain 0 (no pattern 001–011), positions 4–5 increase from 0 to 1 to account for the single occurrence of 100, positions 5–6 go from 1 to 2 for the single occurrence of 101, and the remaining positions stay 2 (no further patterns).

\begin{algorithm}[h]
\caption{Preprocessing }
\label{alg:preprocess}
\begin{algorithmic}[1]
\State \textbf{Input:} Binary matrix \( W \in \{0,1\}^{K \times N} \), factor \( L \in \mathbb{N}\)
\vspace{1mm}
\State \( \{W_1, W_2, \dots, W_{\lceil N/L \rceil}\} \;\leftarrow\; \text{Column Blocking of } W \)
\For{each block \( W_i \)}
    \State \( \sigma_i \;\leftarrow\; \text{Permutation Sequence of } W_i \)
    \State \( \sigma_i' \;\leftarrow\; \text{Reduced Permutation Sequence from } \sigma_i \)
    \State \( s_i \;\leftarrow\; \text{Segmentation Sequence of } \pi_{\sigma_i'}(W_i) \)
\EndFor
\vspace{1mm}
\State \textbf{Return:} \( \{ (\sigma_i', s_i) \mid i = 1, \dots, \lceil N/L \rceil \} \)
\end{algorithmic}
\end{algorithm} 
Algorithm \ref{alg:preprocess} combines the three pre-processing steps. Given fixed ternary weights and block size $L$, preprocessing is applied once, and inference repeatedly reuses the resulting structures for runtime computation. The parameter $L$ is selected empirically per machine; in practice, optimal values are small and strictly below $\log_2 N$ for the theoretical bound to hold. Further, optimal $L$ is decreasing with higher sparsity and increasing with matrix size. Once determined, both preprocessing and inference run with a fixed configuration.

\subsection{Inference}

During inference, given an input matrix $X \in \mathbb{R}^{M \times K}$ with row vectors $\vec{x}_j$ for $j \in \{0, \dots, M-1\}$, we compute the matrix product $X W$ by performing row-wise vector–matrix multiplications $\vec{x}_j \cdot W_i^{[L]}$, where $i \in \{1, \dots, \lceil N/L \rceil\}$, taking advantage of the preprocessed structure of each $W_i^{[L]}$. It involves two steps: segmented sum and tree computation.

\subsubsection{Segmented Sum}

Given $\vec{x}_j \in \mathbb{R}^K$ and $(\sigma_i', s_i)$ for $W_i^{[L]}$, we compute the segmented sum.
\begin{definition}[Segmented Sum]
Let $\sigma_i'$ be the reduced permutation and $s_i$ the segmentation sequence of $W_i^{[L]}$. The \emph{segmented sum} $\vec{u}_i$ is defined as
\[
\vec{u}_i[\ell] = \sum_{r=b_\ell}^{b_{\ell+1}-1} \vec{x}_j[\sigma_i'(r)], \quad \ell \in \{0, \dots , 2^{L}-2\},
\]
where $s_i = \{b_0, \dots, b_{2^L-1}\}$.
\end{definition}

Each entry of $\vec{u}_i$ thus aggregates the contributions of all elements in a row $\vec{x}_j$ sharing the same binary pattern in a column block $W_i^{[L]}$. 

%In Fig.~\ref{fig:example}, the segmented sum for $W_{2,1}$ takes the values of $\vec{x}_j$ from positions 1 and 0 (the reduced permutation sequence) and places them at segmented sum positions 4 and 5, corresponding to patterns 100 and 101 (the 4th and 5th patterns excluding 000). This results in the segmented sum $(0,0,0,0,8,5,0,0)$.

%For example, when the segmentation list is (000: row 0, 001: 0, 010: 0, 011: 0, 100: row 1, 101: row 2, 110: 2, 111: 2), we will omit the 000 pattern and only accumulate the $\vec{x}_j$ data for pattern 100 and 101 according to the Reduced Permuted Sequence (1, 0), which corresponds to (8, 5) of the x vector. Thus, the Segmented Sum is (0, 0, 0, 8, 5, 0, 0).

\subsubsection{Tree Computation} The tree computation can be viewed as a recursive reduction that accumulates contributions from elements according to the binary patterns of their positions.

\begin{definition}[Tree Computation]
Let $L\in \mathbb{N}$ and let $\vec{u}_i$ be the segmented sum vector of length $2^L-1$.  
Define a sequence of vectors $\vec{v}_0, \vec{v}_1, \dots, \vec{v}_{L-1}$ recursively as follows:
\begin{align*}
\vec{v}_0 &:= \vec{u}_i, \\
\vec{v}_{\ell+1} &:= \bigl( \vec{v}_\ell[2], \vec{v}_\ell[4], \dots, \vec{v}_\ell[|\vec{v}_\ell| - 1] \bigr) \text{ for } \ell = \{0,\dots, L-2\}.
\end{align*}
The \emph{tree computation output} is the concatenation of the sums of the even-indexed entries of the vectors in reverse order:\[
\vec{z}_i := \bigl( 
\textstyle\sum_{j=0}^{\big\lfloor\frac{|\vec{v}_{L-1}|}{2}\big\rfloor} \vec{v}_{L-1}[2j], \;\dots,\;
\sum_{j=0}^{\big\lfloor\frac{|\vec{v}_0|}{2}\big\rfloor} \vec{v}_0[2j] 
\bigr).
\]
\end{definition}

The vectors $\vec{v}_0, \vec{v}_1, \dots, \vec{v}_{L-1}$ have odd length, with $\vec{v}_0=\vec{u}_i$ of length $2^L-1$. For $\ell\ge 0$, $\vec{v}_{\ell+1}$ is formed recursively by taking every second element of $\vec{v}_\ell$ from positions $2$ to $|\vec{v}_\ell|-1$, roughly halving the length until $|\vec{v}_{L-1}|=1$. The tree output $\vec{z}_i$ concatenates the even-index sums of each vector in the sequence, i.e. at each level in the tree, in reverse order, as illustrated in Fig.~\ref{fig:example}. The corresponding sums are shown in blue, red, and yellow and the concatenation in reverse order.

\begin{algorithm}[h]
\caption{Inference}
\label{alg:inference}
\begin{algorithmic}[1]
\State \textbf{Input:} $X \in \mathbb{R}^{M \times K}$ and $\{(\sigma_i', s_i)\} \ \text{ for all } W_i^{[L]}$
\For{each row vector $\vec{x}_j \in \mathbb{R}^K$}
    \For{each $(\sigma_i', s_i)$ corresponding to $W_i^{[L]}$}
        \State $\vec{u}_i \gets \text{Segmented Sum of } \vec{x}_j \text{ using } (\sigma_i', s_i)$
        \State $\vec{z}_i \gets \text{Tree Computation of } \vec{u}_i$
    \EndFor
    \State $\vec{Z}_j \gets (\vec{z}_1, \dots, \vec{z}_{\lceil N/L \rceil})$
\EndFor
\State \textbf{Return:} $\vec{Z} \in \mathbb{R}^{M \times N}\text{, where  the $j$-th row is } \vec{Z}_j$
\end{algorithmic}
\end{algorithm}

\subsection{Theoretical Computation Complexity Analysis}
The proposed SSR improves inference efficiency and achieves asymptotically better operational complexity than RSR++ for sparsity $p > 0.5$, as it requires fewer additions and less memory accesses. The preprocessing complexity of SSR remains $O(K \cdot N)$ like in RSR++.

\begin{table}[h!]
\scriptsize
\centering
\caption{Operation counts of SSR and related works.} %\texttt{BitNet.cpp}, RSR++ and
\label{tab: rsr_comparison}
\renewcommand{\arraystretch}{2.4}
\begin{tabular}{l|l|l}
\textbf{\texttt{BitNet.cpp}} & \textbf{RSR++} & \textbf{SSR} \\
\hline
\rule{0pt}{3.5ex}
$
\displaystyle
 \frac{2 M N K}{3}
$
& 
$
\displaystyle
 \frac{ 2 M N K}{\log_2 K}
$
& 
$
\displaystyle
\log_2\!\left(\frac{1}{1 - p}\right) \cdot 
\frac{4 M N \cdot K^{-1/\log_2(1 - p)}}{\log_2 K}
$
\end{tabular}
\end{table}
RSR++ requires, in the worst case, $\frac{2 M N K}{\log_2(K+2)-1}$ operations \cite{RSR_ArXiv_2025}, which for $N \ge 2$ gives the lower bound $2 \cdot \frac{M N K}{\log_2 K}$, as shown in Table~\ref{tab: rsr_comparison}. 
SSR exploits that elements mapped to all-zero bit patterns contribute nothing to the final result, allowing their exclusion from computation. Hence, for each row and block $W_i^{[L]}$, the number of additions drop from $K$ to $K - \mathbb{E}[\texttt{pattern}_0]$, with $\mathbb{E}[\texttt{pattern}_0] = p^L \cdot K$ assuming entries are independently zero with probability $p$. Setting $L = \left\lceil \log_2 K / \log_2\left(\frac{1}{1 - p}\right) \right\rceil$ and following the derivation in \cite{RSR_ArXiv_2025} yields an upper bound on the computational cost, as shown in Table~\ref{tab: rsr_comparison}. The performance difference between RSR++ and SSR arises from both the exponent of $K$ and the constant factor. For $p > 0.5$, we have:
\[
-\log_2\!\left(1-p\right) > 1 \quad \Rightarrow \quad \frac{-1}{\log_2\!\left(1-p\right)} < 1
\]
Hence, SSR achieves strictly better asymptotic performance for $p > 0.5$, coinciding with RSR++ at $p = 0.5$. While SSR has a larger constant factor, the reduced exponent dominates for sufficiently large $K$. Specifically, for sparsity $p \in (0.5,1]$, SSR is more efficient when
\[
K \geq (2c)^{\frac{c}{c-1}} \text{ with} \; c = \log_2\!\left(\frac{1}{1-p}\right).
\]

SSR further reduces computation and memory access. Excluding the all-zero bit pattern shortens the segmented sum array by one entry, saving one addition per row and per column block $W_i^{[L]}$ at each tree level, i.e., at least $M \cdot \frac{N}{L}$ additions. Moreover, the permutation array and segmentation list are shortened, reducing both the number of indices processed and the corresponding memory accesses.

Similar to BitNet \cite{ArXiv_2025_BitNet.cpp} and RSR++ \cite{RSR_ArXiv_2025}, our SSR algorithm is independent of the underlying hardware architecture. Therefore, SSR can be implemented on both CPUs and GPUs. We provide the benchmarking results on CPU only in the next section because of the high engineering effort required.

\section{Evaluation}
We optimize the C++ implementation of SSR with loop unrolling, blocking, AVX2 SIMD intrinsics, and multi-threading to achieve faster speed. We benchmark the SSR CPU implementation with ternary weight matrices at the GEMM level, the layer level, and the model level. The experiments were conducted on a Windows 11 Home 23H2 laptop with an AMD Ryzen 7 8845HS@3.8GHz. The compiler is Microsoft Visual Studio 2022 v143. All experiments are executed for 10 times and we use the average time to draw the graphs. We extend the experiments to Llama Multi-Layer Perceptron (MLP) block and Llama models to systematically evaluate the performance gain of our implementation.

\subsection{Ternary Matrix Multiplication Evaluation} 
\begin{figure}[t]
\centering
\includegraphics[width=\linewidth]{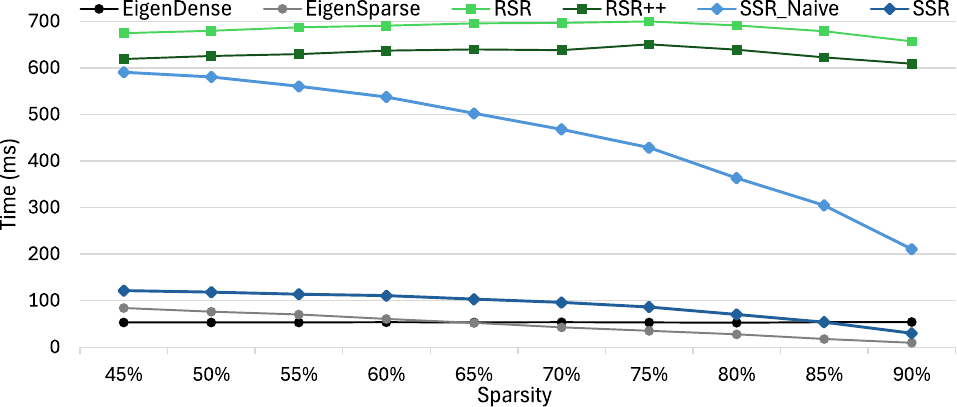}
\caption{Execution time of single-thread ternary GEMM (X: 512$\times$1024, W: 1024$\times$4096, L=6).}
\label{fig-single-thread-GEMM}
\end{figure}

We perform a matrix-level evaluation against RSR and RSR++~\cite{RSR_ArXiv_2025} for the single-threaded implementation, then against more related work for the multi-threaded one. Figure~\ref{fig-single-thread-GEMM} presents the ternary GEMM for both the basic SSR implementation and the AVX2-vectorized version. These experiments are conducted with single-threaded code. The activation matrix has dimensions $512\times 1024$, and the ternary weight matrix has dimensions $1024 \times 4096$. RSR and RSR++ are insensitive to sparsity, as they have almost the same execution time across different sparsity levels. In contrast, our SSR and Eigen Sparse GEMM have shorter execution time with the increasing sparsity consistently. 

RSR++ consistently has 7.6\%-9.1\% improvement over RSR. Our SSR Naive implementation has 2.9\%-189.0\% faster speed over RSR++ thanks to the sparsity. The optimized SSR implementation achieves 4.9-20.2$\times$ speedup over RSR++ when the sparsity ranges from 45\%-95\%. Eigen Dense and Sparse GEMM serve as a very good baseline. Our optimized SSR implementation with AVX2 SIMD achieves similar speed as the Eigen Sparse but cannot outperform it. 

% The results clearly show that the speedup relative to the baseline RSR++ implementation increases with sparsity, reaching a maximum of $10.1\times$ for INT8 precision and $10.2\times$ for float precision with the AVX2-vectorized SSR code. Even at low sparsity levels, the minimum speedups remain at $1.3\times$ (INT8) and $2.3\times$ (float), demonstrating consistent performance improvements across the full sparsity range. In addition, the basic SSR implementation, without any unrolling or vectorization, also consistently outperforms RSR++ across all sparsity levels, achieving maximum speedups of $9.2\times$ for INT8 and $6.0\times$ for float precision. Overall, the results highlight substantial speedups at low sparsity and steadily increasing gains as sparsity grows.

Next, we evaluate the multi-threaded, AVX2-vectorized SSR implementation against established baselines, including PyTorch Dense, PyTorch Sparse CSR and CSC, as well as Eigen Dense and Eigen Sparse implementations. 

We conducted both float and INT8 precision for all experiments. However, the INT8 version of PyTorch and Eigen, including Dense and Sparse GEMM, performed worse than expected. Their INT8 implementations are much slower than their FP32 versions, making these experiment results unusable. For example, the Eigen Sparse GEMM INT8 is 2.2-4.5$\times$ slower than its FP32 version, while others may be even slower than PyTorch Sparse CSR. Therefore, though our INT8 speedup is relatively better than the speedups in FP32, we exclude these INT8 benchmarking results and only provide Eigen Sparse GEMM INT8 for reference. 

\begin{figure}[b]
\centering
\includegraphics[width=\linewidth]{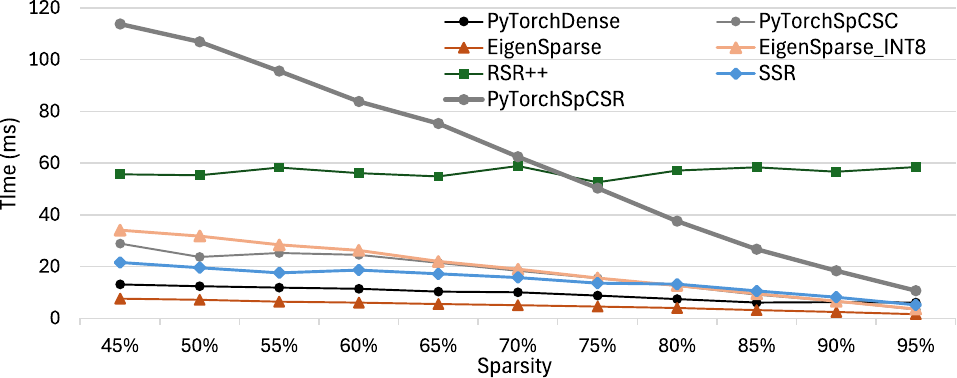}
\caption{Execution time of multi-thread ternary GEMM (X: 256$\times$1024, W: 1024$\times$4096, L=4).}
\label{fig-GEMM-1kx4k}
\end{figure}

Figure~\ref{fig-GEMM-1kx4k} shows the ternary GEMM time with a small weight matrix of size $1024 \times 4096$. PyTorch Dense GEMM is a strong baseline as it has been optimized by many engineers and widely applied in the industry. However, the PyTorch Sparse CSR (Compressed Sparse Row) performs worse than other sparse methods. It only outperforms RSR++ with 75\% or higher sparsity. Our SSR achieves 14.0\%-43.8\% faster speed than PyTorch Sparse CSC (Compressed Sparse Column) with 45\%-75\% sparsity, but is up to 35\% slower than it with very high sparsity. Similarly, SSR is 14.7\%-62.2\% faster than Eigen Sparse INT8 GEMM when the sparsity is lower than 80\%. As RSR++ has similar execution time for different sparsity levels, our SSR achieves 2.6-11.3$\times$ speedup over RSR++.

% SSR\_vec delivers good performance in INT8 precision, achieving speedups from roughly $22\times$ at low sparsity to as high as $88\times$ at high sparsity levels, consistently outperforming competing algorithms. Its performance scales exponentially with sparsity, reflecting exploitation of sparse patterns in an exponential tree computation. In contrast, for floating-point precision, SSR\_vec achieves more modest speedups of $0.3$--$1.5\times$, although its performance improves with increasing sparsity and at high sparsity levels can surpass PyTorch dense and sparse implementations. Hence, SSR\_vec is particularly well-suited for quantized weight networks with INT8 activation values.

\begin{figure}[t]
\centering
\includegraphics[width=\linewidth]{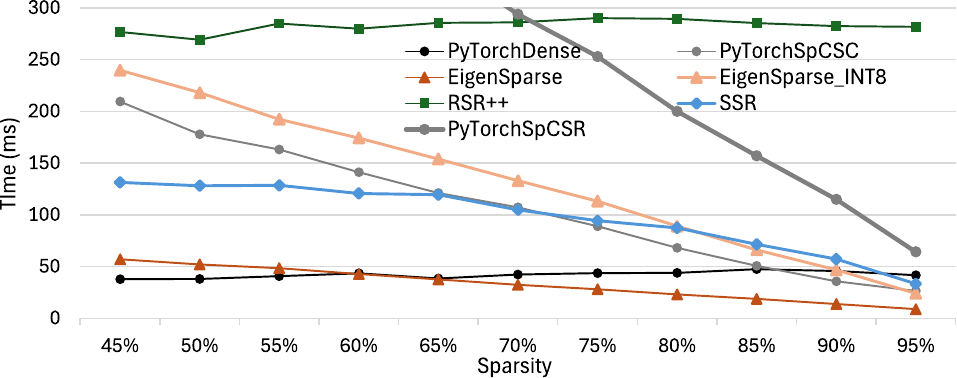}
\caption{Execution time of multi-thread ternary GEMM (X: 512$\times$2048, W: 2048$\times$8192, L=6).}
\label{fig-GEMM-2kx8k}
\end{figure}

Fig.~\ref{fig-GEMM-2kx8k} shows ternary GEMM with a larger weight matrix, zoomed in to highlight performance trends. Eigen Sparse GEMM tends to perform better on smaller matrices, as its relative speed compared with PyTorch Dense GEMM shifts slower compared with Fig.~\ref{fig-GEMM-1kx4k}. The trend is similar to the previous graph. SSR is up to 59.4\% and 82.5\% faster than PyTorch Sparse CSC and Eigen Sparse INT8 with 70\% or lower sparsity, but becomes slower at higher sparsity. SSR achieves 2.1–8.4$\times$ speedup over RSR++ for 45–95\% sparsity, due to large matrix sizes (N$>>$2) and lower memory footprint, which we evaluate later in layer-level benchmarking.

The performance curves in Fig. \ref{fig-GEMM-1kx4k} and Fig. \ref{fig-GEMM-2kx8k} show that our SSR seems to do not utilize the sparsity as good as other methods. The reason is that these experiments use a fixed $L$ factor. Based on our experiments, the optimal $L$ value changes with both the matrix size and the sparsity level. Therefore, dynamically selecting the $L$ value based on a performance profiling can overcome this limitation.

\subsection{Layer-Level Evaluation} 
\begin{figure}[t]
\centering
\includegraphics[width=\linewidth]{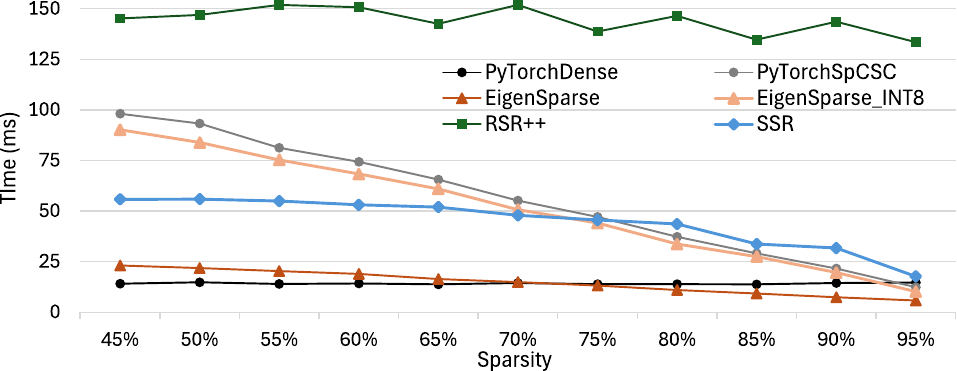}
\caption{Execution time of one small Llama MLP block (X: 256$\times$1024, up\_proj: 1024$\times$4096, L=6).}
\label{fig-MLP-1kx4k}
\end{figure}

\begin{figure}[b]
\centering
\includegraphics[width=\linewidth]{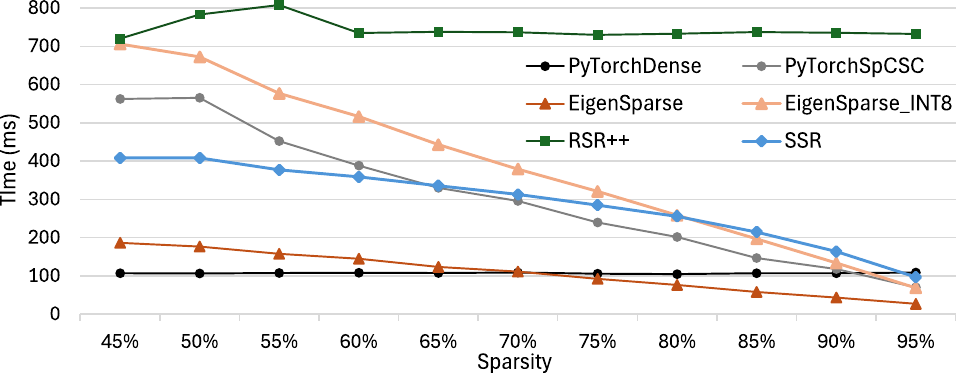}
\caption{Execution time of one Llama-3 1B MLP block (X: 512$\times$2048, up\_proj: 2048$\times$8192, L=6).}
\label{fig-MLP-2kx8k}
\end{figure}

We evaluate the ternary GEMM performance at Llama-3 Multi-Layer Perceptron blocks. As one MLP block contains one up projection (e.g., up\_proj in Fig.~\ref{fig-MLP-1kx4k}: 1024$\times$4096), one gate projection (1024$\times$4096), and one down projection (4096$\times$1024), it reflects how the GEMM performance contributes to the layer-level performance. The activation function like GeLU or SiLU are omitted here to only evaluate the ternary GEMMs. The sizes of other projections changes with the up projection accordingly in Fig.~\ref{fig-MLP-2kx8k}. 

%(see ). SSR and SSR\_vec achieve dramatically superior performance over RSR++, delivering 10--20$\times$ better speedups across all sparsity levels in both float and INT8 precision. When comparing SSR variants against RSR++, the performance advantage reaches up to 41$\times$ speedup in float and 38$\times$ in INT8 for high sparsities, demonstrating the substantial computational efficiency gains of the SSR approach, which becomes more pronounced with high sparsity. 

Fig.~\ref{fig-MLP-1kx4k} presents the execution time of one small MLP block. Though the MLP block execution time is about 2.5-3$\times$ as long as one single GEMM across different methods, it shared similar trend as the GEMM evaluation results. SSR is up to 75.8\% faster but up to 31.9\% slower than PyTorch CSC, which shows a better scaling curve across the sparsity than the single GEMM case. The crossing point is still below 80\% sparsity. SSR has 2.6-7.5$\times$ speedup over RSR++ at this MLP block. 

Fig.~\ref{fig-MLP-2kx8k} presents the execution time of one Llama-3 1B MLP block, which is about 8$\times$ as large as the small MLP block. Similar to the GEMM benchmarking, the Eigen Sparse INT8 has a worse speed than the small case, and is 13.7-34.2\% slower than the PyTorch CSC. SSR is up to 38.5\% faster and 31.5\% slower than PyTorch CSC. Compared with PyTorch Dense, it achieves faster speed at 95\% sparsity, which is the same as PyTorch CSC and Eigen Sparse INT8. SSR achieves 1.8-7.5$\times$ speedup over RSR++.

\begin{table}[h]
\caption{Memory usage of one Llama-3 1B MLP block}
\label{tab-llama-3-MLP}
\centering
\begin{tabular}{r|r|r}
\hline
      & After Initialization & During Inference \\ \hline
RSR++ & 129-146 MB           & 177-259 MB       \\ \hline
SSR   & 122-124 MB           & 160-213 MB       \\ \hline
\end{tabular}
\end{table}

In addition to the speed analysis, we provide the memory usage of the MLP block in Table \ref{tab-llama-3-MLP}. The memory usage after initialization shows the data format size that store the ternarized weights. Compared with RSR++, SSR reduces 5.4-15.1\% of memory thanks to the efficient data format that removes redundant all-zero indices. The memory usage during inference shows the dynamic memory in execution. As SSR can utilize the sparsity to skip computation with zero weights, we reduces the memory usage by 9.6-17.8\% compared with RSR++. Such a smaller memory footprint not only enhances performance but also lowers energy consumption due to fewer memory accesses

%When executing Llama-3 MLP blocks.

\subsection{End-to-End Evaluation} 
We report the Time-To-First-Token latency of a ternary llama-3.1 1B model in Table~\ref{tab:ttft-llama31-1b}. The MLP blocks of the llama model are replaced with corresponding ternary MLP blocks, and the activations like SiLU are added back to these MLP blocks. The attention part are implemented using libTorch C++ APIs with dedicated PyTorch native Multi-Head-Attention functions to maximize its efficiency. We set the sequence length to 512 to reflect a relative long user query to that language model.
% reports the memory usage of a single LLaMA-3 1B MLP block under RSR++ and SSR. During initialization, both methods require comparable memory, with RSR++ using 129–146 MB and SSR slightly less at 122–124 MB. During inference, the gap widens: RSR++ consumes 177–259 MB, while SSR reduces this to 160–213 MB. 

The Time-To-First-Token latency of RSR++ ternary Llama is very stable across the tested sparsity levels, as it does not exploit sparsity to remove redundant computations. In contrast, our SSR achieves lower latency and faster inference, with a 1.84$\times$ speedup at 90\% sparsity compared to at 50\%. This relative speedup is smaller than the 5$\times$ theoretical speedup because the attention layers are not ternarized. Compared with RSR++, SSR achieves a 3.45--6.34$\times$ speedup for long queries with 512 tokens, with further improvements possible by ternarizing Multi-Head Attention and dynamically selecting $L$ instead of using a fixed value.

%The Time-To-First-Token latency of RSR++ ternary Llama is very stable across the tested sparsity levels, as it does not utilize the sparsity to reduce redundant computations. In contrast, our SSR ternary Llama achieves faster speed with lower latency. It achieves 1.84$\times$ speedup with 90\% of sparsity over inference with 50\% of sparsity. This relative speedup across sparsity is smaller than the 5$\times$ theoretical speedup (50\% density over 10\% density) because the attention part is not ternarized. It may achieve faster speed if we ternarize the Multi-Head-Attention computation. Compared with RSR++, SSR achieves 3.45-6.34$\times$ speedup on the long query with 512 tokens. Note that this experiment was conducted with fixed k value. A wrapper that dynamically selects the best k may further boost the performance.

At the full model level, RSR++ requires up to 3113~MB to execute the ternary LLaMA-3.1 1B model with a sequence length of 512, whereas SSR reduces this to 2960~MB, achieving a consistent memory overhead reduction as the GEMM-level and layer-level benchmarking results. The 4.9\% of memory reduction percentage is smaller than the MLP blocks due to the addition of the attention functions. 
%The latency benefits shown in Table~\ref{tab:ttft-llama31-1b} are even more pronounced: SSR delivers a 3.5$\times$ speedup in time-to-first-token at 50\% sparsity, scaling to a 6.3$\times$ speedup at 90\% sparsity. SSR outperforms RSR++ in both absolute performance and sparsity scaling.

%RSR++ uses up to 3113 MB of memory when executing ternary LLama-3.1 1B model. The token length is 512. SSR uses up to 2960 MB of memory \ref{tab:ttft-llama-3 MLP}. For time to first token:: SSR achieves a 3.5$\times$ speedup at 50\% sparsity, which further improves to 6.3$\times$ at 90\% sparsity compared to RSR++.  
%This demonstrates that SSR not only outperforms RSR++ in absolute latency but also benefits more strongly as sparsity increases. \ref{tab:ttft-llama31-1b}

\begin{table}[t]
\centering
\caption{Time-To-First-Token of one Llama-3.1 1B model (Sequence Length = 512)}
\label{tab:ttft-llama31-1b}
\begin{tabular}{r|r|r|r}
\hline
Sparsity & 50\% & 70\% & 90\% \\ \hline
RSR++    & 36.23 s & 36.30 s & 36.19 s \\ \hline
SSR      & 10.49 s &  8.80 s &  5.71 s \\ \hline
\end{tabular}
\end{table}

%\subsection{Limitations and Future Work}

\section{Conclusion}

In this paper, we present Sparse Segment Reduction (SSR) as an optimized ternary GEMM method for ternary LLMs and general TWNs. Related works BitNet \cite{ArXiv_2025_BitNet.cpp} RSR++ \cite{RSR_ArXiv_2025} reduce the complexity of ternary GEMM but overlook zero values. Therefore, we leverage the zero weights in the ternary GEMM to reduce storage and eliminate redundant computations. SSR features a specialized compact data format that reduces up to 15\% memory size than RSR++ and a sparse GEMM algorithm that achieves another log-scale improvement on the computation complexity. We present the pre-processing algorithm for the data format and the inference algorithm with tree-reduction in mathematic equations. We also mathematically show the computation complexity of SSR and the boundary conditions. We implement the proposed SSR in C++ and optimize the code with standard optimizations like loop unrolling, AVX2 SIMD, and multi-threading. Evaluation results show that SSR achieves a speed up to 75.8\% faster than PyTorch Sparse CSC and a speedup of 2.1-11.3$\times$ over RSR++ in the ternary GEMM. SSR achieves 3.5-6.3$\times$ end-to-end speedup over RSR++ on the ternary Llama-3 1B model with 4.9\% of total memory reduction. 

We hope that this work can facilitate the adoption of ternary LLMs and general TWNs. The limitation of this work is the lack of implementation for GPUs and more LLM architectures. This points to future work in GPU implementation and inference demonstration on open-weight ternary LLMs. Given the strong performance of related works BitNet and RSR++ on GPUs, SSR is also expected to yield further improvements with GPU-specific optimizations. 

%We have presented an optimized approach for ternary weight network inference on AVX2-enabled edge devices. Our algorithm leverages naturally occurring sparsity to significantly improve upon RSR++, delivering both theoretical runtime benefits and practical speedups of up to $10\times$ across all sparsity levels. Implemented in C++, it consistently outperforms state-of-the-art methods in both single-threaded and multi-threaded settings, demonstrating its effectiveness for efficient inference on resource-constrained hardware. 

%The MLP benchmarks demonstrate that SSR and SSR\_vec achieve remarkable performance gains over RSR++, delivering 10--20$\times$ speedup across a wide range of sparsity levels and up to 41$\times$ in float and 38$\times$ in INT8 at high sparsity levels.

\bibliographystyle{IEEEtran}
\bibliography{reference}

\end{document}